\documentclass[letterpaper, 10 pt, conference]{ieeeconf}  

\usepackage{amsmath,amsfonts}
\usepackage{algorithmic}
\usepackage{algorithm}
\usepackage{array}
\usepackage[caption=false,font=small,labelfont=sf,textfont=sf]{subfig}
\usepackage{textcomp}
\usepackage{stfloats}
\usepackage{url}
\usepackage{verbatim}
\usepackage{graphicx}
\usepackage{cite}
\usepackage[OT1]{fontenc}
\usepackage[colorlinks, linkcolor=blue, anchorcolor=blue, citecolor=blue]{hyperref}
\usepackage{amssymb}
\usepackage{multirow}
\usepackage{marvosym}

\graphicspath{{Figures/}}

\IEEEoverridecommandlockouts                              

\title{\LARGE \bf DTRT: Enhancing Human Intent Estimation and Role Allocation for Physical Human-Robot Collaboration}

\author{Haotian Liu$^{1}$, Yuchuang Tong$^{1,}\textsuperscript{\Letter}$, Zhengtao Zhang$^{1,2,}\textsuperscript{\Letter}$
\thanks{This work was supported by the National Natural Science Foundation of China (62303457, U21A20482), Project funded by China Postdoctoral Science Foundation (2023M733737), and the Beijing Municipal Natural Science Foundation, China (4252053).}
\thanks{$^{1}$ CAS Engineering Laboratory for Intelligent Industrial Vision, Institute of Automation, Chinese Academy of Sciences, Beijing 100190, China, and also with the School of Artificial Intelligence, University of Chinese Academy of Sciences, Beijing 100049, China. {\tt\small liuhaotian2021, yuchuang.tong, zhengtao.zhang@ia.ac.cn}}
\thanks{$^{2}$ Beijing Zhongke Huiling Robot Technology Co., LTD, Beijing 100192, China.}
\thanks{$\textsuperscript{\Letter}$ Corresponding author.}}

\begin{document}

\maketitle
\thispagestyle{empty}
\pagestyle{empty}

\begin{abstract}

    In physical Human-Robot Collaboration (pHRC), accurate human intent estimation and rational human-robot role allocation are crucial for safe and efficient assistance. Existing methods that rely on short-term motion data for intention estimation lack multi-step prediction capabilities, hindering their ability to sense intent changes and adjust human-robot assignments autonomously, resulting in potential discrepancies. To address these issues, we propose a Dual Transformer-based Robot Trajectron (DTRT) featuring a hierarchical architecture, which harnesses human-guided motion and force data to rapidly capture human intent changes, enabling accurate trajectory predictions and dynamic robot behavior adjustments for effective collaboration. Specifically, human intent estimation in DTRT uses two Transformer-based Conditional Variational Autoencoders (CVAEs), incorporating robot motion data in obstacle-free case with human-guided trajectory and force for obstacle avoidance. Additionally, Differential Cooperative Game Theory (DCGT) is employed to synthesize predictions based on human-applied forces, ensuring robot behavior align with human intention. Compared to state-of-the-art (SOTA) methods, DTRT incorporates human dynamics into long-term prediction, providing an accurate understanding of intention and enabling rational role allocation, achieving robot autonomy and maneuverability. Experiments demonstrate DTRT's accurate intent estimation and superior collaboration performance.

\end{abstract}

\section{Introduction}

Physical Human-Robot Collaboration (pHRC) is essential in manufacturing\cite{xingImpedanceLearningHumanGuided2023}, healthcare\cite{10309357}, and services\cite{yuAdaptiveConstrainedImpedanceControl2022}. Effective strategies are needed to ensure robots seamlessly collaborate with humans, accurately estimate intentions, dynamically adjust behaviors, and assist humans with minimal effort\cite{raviProactiveSafeHumanRobot2024}. Therefore, accurate human intent estimation and rational human-robot role allocation are critical challenges for improving pHRC performance.

Accurate prediction of future trajectories based on human intent is crucial for efficient robot assistance and ensuring safe pHRC\cite{mengHierarchicalHumanMotion2024}. In complex environments with potential hazards, such as unknown obstacles for the robot, rapid changes in human intent pose significant challenges for intent estimation. Current methods primarily rely on short-term motion data, such as position and velocity, which limits the ability to detect changes in human intentions and affects the accuracy and safety of predictions\cite{liuFollowForceHaptic2023, zhangOnlineHumanDynamic2023}. Moreover, short-term data reduces the effectiveness of intent estimation in long-term collaborations. Therefore, investigating long-term prediction methods that integrate human-applied forces in pHRC is essential for achieving accurate intent estimation.

Moreover, human-robot role allocation involves a sophisticated mechanism of assigning task control between human and robot. This process coordinates the human-robot relationship in real-time to reduce disagreements and improve robot assistance level. Existing methods primarily rely on impedance/admittance control, where roles are determined by modifying model parameters\cite{8616062, luoPhysicalHumanRobot2024}. Among the various methods, game theory-based controllers\cite{franceschiAdaptiveImpedanceController2022} simulate the collaboration process between multi-players, framing the objective as minimizing a cost function to achieve optimal pHRC. Nonetheless, ensuring that robot behaviors align with human intentions while maintaining autonomy and maneuverability remains a significant challenge.

\begin{figure}[!t] 
    \centering
    \includegraphics[width=0.47\textwidth]{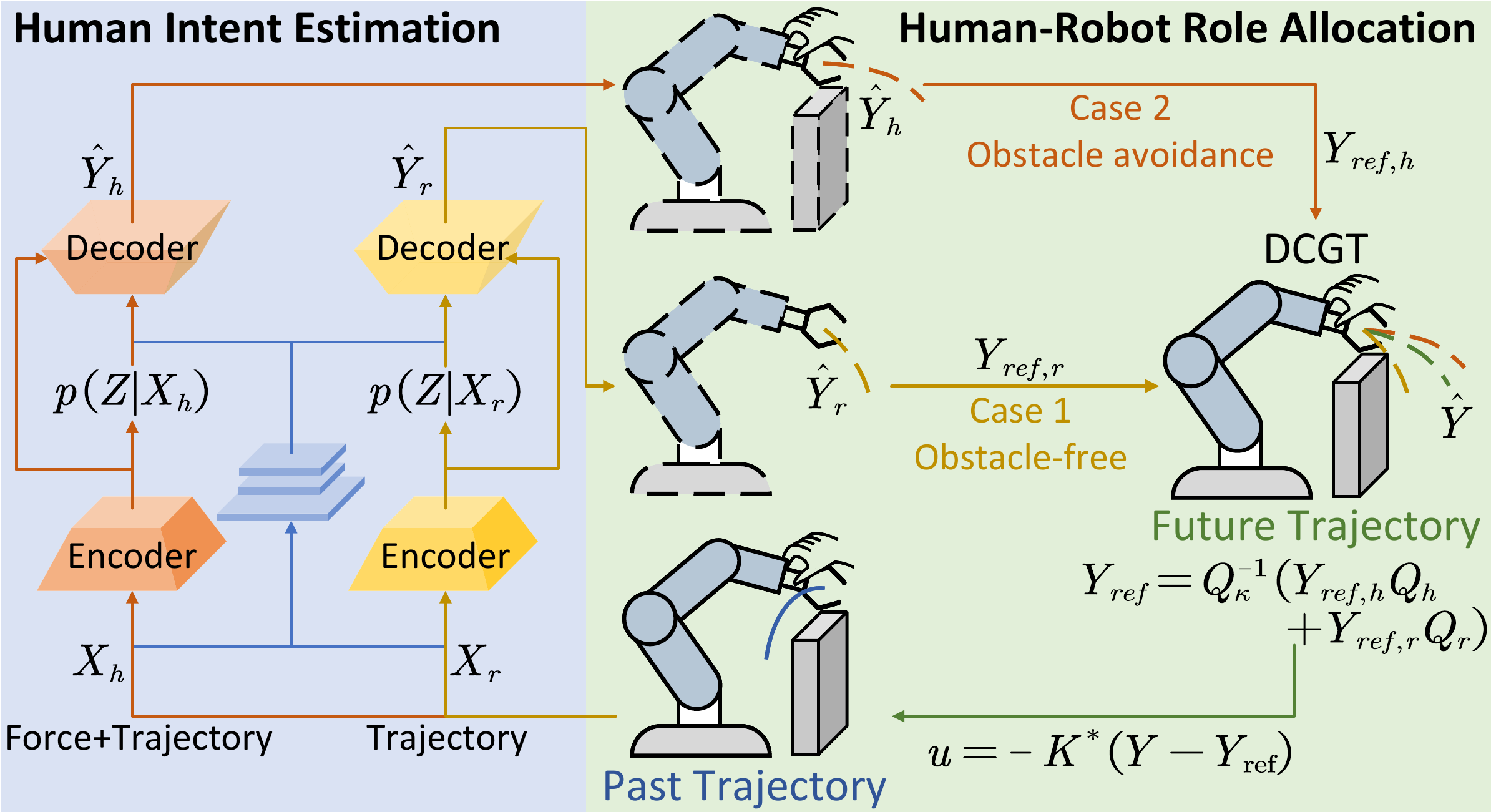}
    \caption{Dual Transformer-based robot trajectron for human intent estimation and role allocation.}
    \label{fig:2.2}
\end{figure}

For these purposes, we propose a Dual Transformer-based Robot Trajectron (DTRT), as shown in Fig. \ref{fig:2.2}. DTRT employs a hierarchical architecture with two Transformer-based Conditional Variational Autoencoders (CVAEs) and introduces force into long-term prediction, enabling rational human-robot role allocation and reducing discrepancies based on Differential Cooperative Game Theory (DCGT). Compared to existing methods \cite{songRobotTrajectronTrajectory2024,cremerModelFreeOnlineNeuroadaptive2020, maHumanRobotCollaboration2024,franceschiLearningHumanMotion2023}, DTRT offers an accurate understanding of intentions by integrating human dynamics into long-term prediction, thereby enhancing estimate accuracy when human intent changes. Furthermore, it autonomously allocates the roles of leader and follower, maintaining robot autonomy in obstacle-free case and improving assistance during obstacle avoidance. DTRT demonstrates remarkable performance on several pHRC metrics, making it highly suitable for pHRC applications. The main contributions of this paper are as follows.

\begin{itemize}
    \item [1)] The proposed DTRT integrates human intent estimation with role allocation to detect intent changes and reduce discrepancies, effectively enhancing collaborative performance in complex and hazard-prone environments.
    
    \item [2)] The hierarchical structure of human intent estimation in DTRT enables simultaneous processing of motion and force in pHRC, improving prediction accuracy and providing an accurate understanding of human intentions.

    \item [3)] DCGT-based role allocation enables adaptive leader switching based on human-applied force, ensuring robot behavior aligns with human intentions and reducing discrepancies while preserving robot autonomy.
    


\end{itemize}


\section{Related Work}




{\noindent \bf Physical Human-Robot Collaboration.} PHRC allows robots to better understand and respond to human needs through physical coupling between humans and robots\cite{shao2024constraintaware}. Current research on pHRC primarily focus on accurately recognizing and predicting human intentions using machine learning methods, as well as ensuring compliance during collaborations through impedance/admittance control\cite{sharifiImpedanceLearningBasedAdaptive2022, pengNeuralControlHuman2024, wangRoleDynamicAssignment2024}. However, improving robot maneuverability and reducing human-robot discrepancies in complex environments remains a significant challenge \cite{xingDynamicMotionPrimitivesBased2024}. Key issues include improving human intent estimation accuracy and robot assistance level while ensuring safety, particularly in hazardous situations.

{\noindent \bf Human Intent Estimation.} The accuracy of human intent estimation has a significant impact on collaboration performance. 
Traditional human intent estimation methods use high-precision force sensors to measure external human-applied forces, reflecting the direction of human motion\cite{Huang2005SharedNC, 6054057, yuAdaptiveConstrainedImpedanceControl2022}. Alternatively, human intent can be recognized by decoding neural signals such as electromyography or electroencephalography\cite{6943678, 6683036}. With the development of neural networks, Cremer et al.\cite{cremerModelFreeOnlineNeuroadaptive2020} employed a two-layer neural network, and Ma et al.\cite{maHumanRobotCollaboration2024} incorporated a Bayesian Neural Network (BNN) with it for intent estimation. However, these short-term prediction methods underutilize prior information, and are limited to next-step predictions, leading to suboptimal local decisions. For long-term predictions, Recurrent Neural Networks (RNNs) and Transformers have garnered considerable attention due to their ability to handle sequential data \cite{gaoHybridRecurrentNeural2023, dominguez-vidalExploringTransformersVisual2024, fuscoTransformerBasedPredictionHuman2024}. These methods take contextual information into account, allowing the information to persist, thereby improving estimation accuracy since previous motion states may reflect underlying intentions\cite{franceschiLearningHumanMotion2023}. Moreover, to capture the multi-modal characteristics of trajectories, Song et al.\cite{songRobotTrajectronTrajectory2024} proposed Robot Trajectron (RT) that generates probabilistic representations of predicted trajectories based on historical motion, and samples future trajectories.

RT combines LSTM with CVAE framework, demonstrating its effectiveness in teleoperation tasks. However, LSTM's limitations in handling long-range dependencies and parallel processing capabilities result in less efficiency and accuracy. Additionally, RT focus on motion information limits its ability to detect human intent changes, which hinders robot from adjusting its behavior to follow human. To address this limitation, we introduce a hierarchical architecture that separately manages standard and emergent situations. By incorporating force, robot can gain an accurate understanding of human intentions, enhancing its assistance level.

{\noindent \bf Human-Robot Role Allocation.} In pHRC, the robot's role must seamlessly transition between leader and follower to ensure task execution\cite{7097058}. To achieve this, the role allocation in shared control between humans and robots needs to be properly designed to assign a degree of leadership to the robot while aligning with human intentions, which helps human accomplish task with minimal effort\cite{xingFuzzyLogicBasedArbitration2024}. As discussed in \cite{7139983}, Game Theory (GT) provides an effective means of analyzing complex interaction behaviors involving players. GT-based methods model the interactions between players by formulating objectives as cost function minimization problems, determining optimal behaviors. \cite{7548305, 7989336, liDifferentialGameTheory2019} explored Non-Cooperative Game Theory (NCGT) to derive optimal behaviors. However, Nash equilibria, the solutions to non-cooperative games, are often not Pareto optimal. To address this issue, Franceschi et al. \cite{franceschiAdaptiveImpedanceController2022} investigated DCGT to describe collaborative models and seek solutions. The cooperative description of tasks enables seamless role allocation, facilitating optimal co-manipulation of objects. Subsequently, \cite{franceschiHumanRobotRole2023} provided further discussion on DCGT and NCGT. DCGT demonstrates strong capabilities in continuously managing leader-follower transition, making it effective for describing human-robot collaboration tasks.

Based on the advantages of DCGT, we synthesize the results of human intent estimation. Furthermore, for safe pHRC in complex scenarios with potential hazards, we incorporate human-applied force to construct the role allocation coefficient. Consequently, DTRT can dynamically adjust the pHRC relationship, ensuring safe and efficient completion of tasks while aligning with human intentions.

\section{Dual Transformer-Based Robot Trajectron}

This section presents DTRT for human intent estimation and role allocation in pHRC, capable of effectively mitigating potential hazards and enhancing robot assistance level.

\subsection{Problem Formulation} \label{ss2}

In this paper, we explore scenarios where humans and robots collaborate in environments with potential hazards, such as obstacles that the robot cannot autonomously avoid. In such situations, human intent changes to avoid danger, prompting robot to adjust its behavior accordingly. We split these scenarios into two distinct cases, as shown in Fig. \ref{fig:2.2}.

{\noindent \bf Case 1: Obstacle-Free.} In this case, human-applied force $f_h$ is close to 0, with the robot autonomously assuming leader. The position and velocity of end-effector are $x_r, \dot{x}_r$. Robot as the leader avoids measurement errors and environmental disturbances in human-applied force.

{\noindent \bf Case 2: Obstacle Avoidance.} Here, human guides robot to adjust its trajectory to avoid hazards. The position and velocity are $x_h, \dot{x}_h$. Thus, human takes on the leading role and robot provides assistance to human. By incorporating human-applied force $f_h$ in prediction, robot can better understand human intentions and avoid hazards in the environment.


Assuming current time is $T_{\mathrm{now}}$ and observable time step is $T_{\mathrm{obs}}$, we represent past data as $X_r=[x_r^t,\dot{x}_r^t]_{t=T_{\mathrm{now}}-T_{\mathrm{obs}}}^{T_{\mathrm{now}}}$ and $X_h=[x_h^t,\dot{x}_h^t,f_h^t]_{t=T_{\mathrm{now}}-T_{\mathrm{obs}}}^{T_{\mathrm{now}}}$. Similarly, for future time step $T_{\mathrm{fut}}$, future data are $Y_r=[x_r^t, \dot{x}_r^t]_{t=T_{\mathrm{now}}}^{T_{\mathrm{now}}+T_{\mathrm{fut}}}$ and $Y_h=[x_h^t, \dot{x}_h^t]_{t=T_{\mathrm{now}}}^{T_{\mathrm{now}}+T_{\mathrm{fut}}}$. Finally, we synthesize the predictions $\hat{Y}_r, \hat{Y}_h$ based on DCGT. This involves adjusting weights according to human-applied force, achieving a weighted composition of $\hat{Y}_r$ and $\hat{Y}_h$, ensuring robot's optimal behavior.

\subsection{Human Intent Estimation}

\begin{figure}[!t] 
    \centering
    \includegraphics[width=0.485\textwidth]{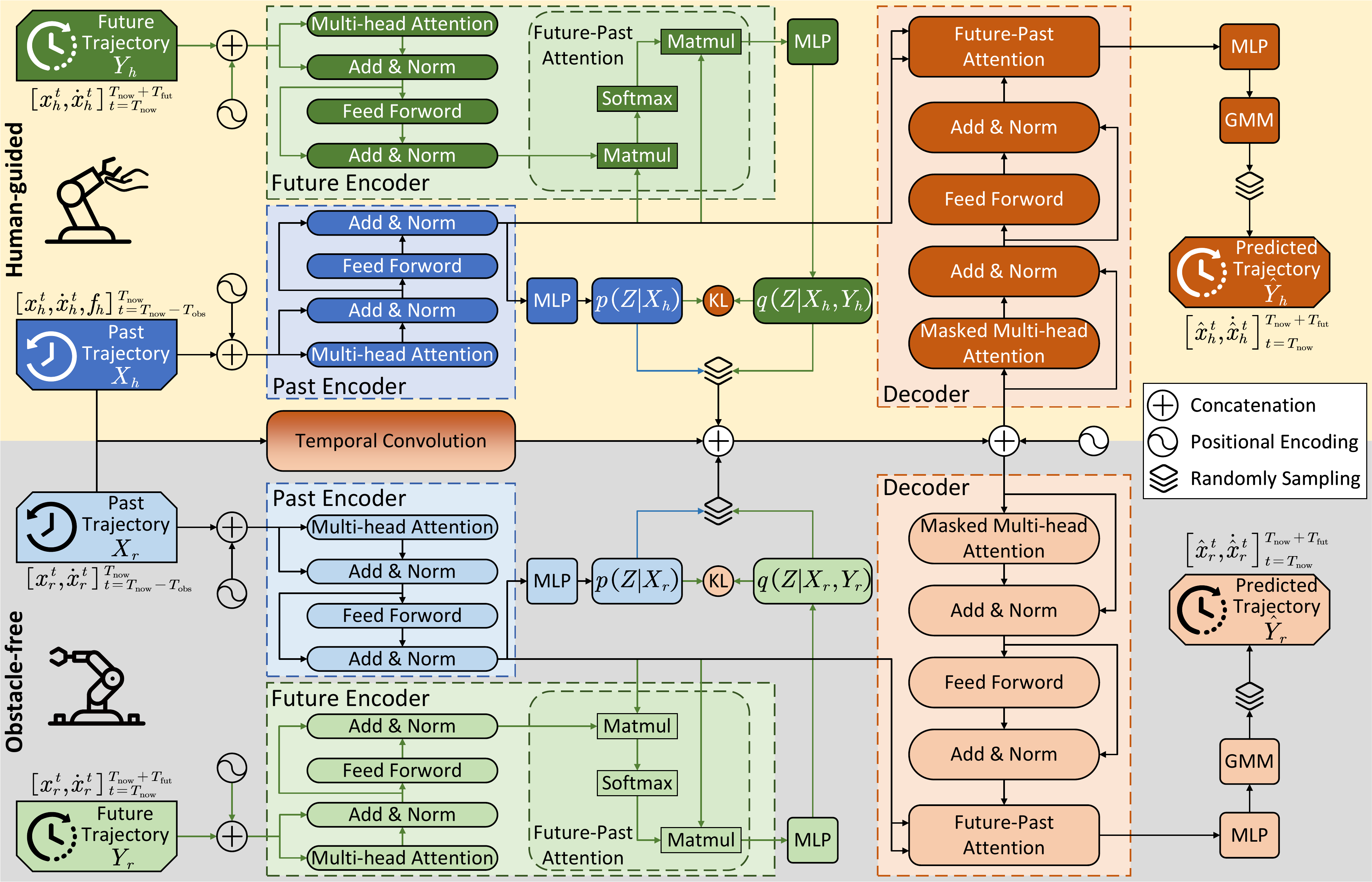}
    \caption{The structure of human intent estimation in DTRT. The green lines denote the train-only operations, while the blue lines denote the predict-only operations.}
    \label{fig:2.1}
\end{figure}

To enhance robot assistance in pHRC, it is essential to estimate human intentions and predict future motions. The human limb dynamics can be defined as
\begin{equation}
    D_h(\dot{x}_h-\dot{x})+K_h(x_h-x)=f_h
    \label{2.13}
\end{equation}
where $D_h$ and $K_h\in \mathcal{R}^{6\times 6}$ are the damping and stiffness matrices of human. Therefore, the desired human motion can be modeled as a nonlinear function $\mathcal{F} (x,\dot{x},f_h)$. This complex human dynamics can be learned by neural networks.

To predict the future trajectory, we model it as a conditional distribution of the past by learning the multi-modal probability distribution function $p(Y|X)$. We employ the CVAE framework and introduce Gaussian latent variable $Z$. Thus, the latent distribution can be expressed as
\begin{equation}
    p(Y|X)=\int p(Y|X,Z)p(Z|X)dZ
    \label{2.1}
\end{equation}

Next, we encode the probability distribution by Transformer and Linear layer, and optimize it by minimizing loss $L$, which consists of the weighted negative evidence-based lower bound (ELBO):
\begin{equation}
    \begin{aligned}
        L 
        &=\kappa D_{KL}(q(Z|X,Y)||p(Z|X))\\
        &\quad- \beta E_{Z\backsim q(Z|X,Y)}[\log p(Y|X,Z)]
    \end{aligned}
    \label{2.2}
\end{equation}







The architecture of human intent estimation in DTRT is shown in Fig. \ref{fig:2.1}. Before learning features, we use the positional encoding method of Transformer to represent temporal information of data. This encoding is concatenated with the data input embeddings and fed into the estimator.

In DTRT, we learn the temporal dependencies of data via multi-head self-attention. Specifically, given the data of end-effector over $T$ time steps, we first extract the features from the input and perform linear projections to obtain the keys $K$, queries $Q$, and values $V$. These mapped keys, queries, and values are fed into self-attention to learn the features as
\begin{equation}
    F=\mathrm{Softmax}(\frac{QK^\mathrm{T}}{\sqrt{d_\mathrm{k}}})V
    \label{2.3}
\end{equation}

The past encoder encodes past data into a latent distribution by employing multi-head self-attention combined with standard feedforward and norm layers to capture the features of the past. Then, it is fed into a multi-layer perceptron (MLP) to generate the Gaussian parameters of $p(Z|X)$.

Similarly, future encoder encodes future data into a latent distribution to capture features. Additionally, we incorporate a future-past attention module to update the features by learning the relationship between past and future data. This is accomplished by integrating the past data features $K$ and $V$ with the future data features $Q$, following the standard feedforward and norm layers. Finally, it is fed into an MLP to generate the Gaussian parameters of $q(Z|X,Y)$.

In the decoder, we combine sampled result $Z$ with temporal convolution result of past data as input. During training, $Z$ is obtained from $q(Z|X,Y)$ generated by the future encoder. During testing, it is obtained from $p(Z|X)$ generated by the past encoder. The structure of decoder is similar to the future encoder. Additionally, we apply proper masks to prevent model from attending to subsequent data when predicting.

Finally, the output serves as parameters of a Gaussian Mixture Model (GMM), from which we sample the prediction
\begin{equation}
    \begin{aligned}
        &Z_{\mathrm{best}}=\underset{Z}{\operatorname*{argmax}}\ p_{\theta}(Z|X),\\
        &\hat{Y}=\underset{Y}{\operatorname*{argmax}}\ p(Y|X,Z_{\mathrm{best}})
    \end{aligned}
    \label{2.4}
\end{equation}


In this process, we introduce a hierarchical architecture for long-term prediction of future trajectories $\hat{Y}_h,\hat{Y}_r$. Two branches respectively handle human-guided obstacle avoidance data and obstacle-free trajectories, which are integrated in role allocation. The incorporation of Transformer significantly enhances prediction accuracy of DTRT. Compared to SOTA methods\cite{songRobotTrajectronTrajectory2024, franceschiLearningHumanMotion2023, 9412190}, human intent estimation in DTRT exhibits a notable advantage, particularly when trajectories are deviated by sudden changes in human intentions.

\subsection{DCGT-based Role Allocation} 

Next, we introduce DCGT to rationally allocate human-robot roles and reduce discrepancies based on human intent estimation. The reference model of robot can be designed as
\begin{equation}
    M\ddot{e}+D\dot{e}+Ke=f_h+f_r
    \label{2.5}
\end{equation}
where $M, D, K\in\mathcal{R}^{6\times 6}$ are the desired inertia matrices, Coriolis/centrifugal force matrices, and gravity, respectively. $e=x_{ref}-x$ is the tracking error, in which $x_{ref}$ is the reference trajectory. $x, \dot{x}, \ddot{x}\in \mathcal{R}^{6}$ are the actual position, velocity, and acceleration of end-effector, $f_r\in \mathcal{R}^{6}$ is the virtual robot effort and $f_h\in \mathcal{R}^{6}$ is the human-applied force. 

First, we introduce a state variable $Y=[x,\dot{x}]^T$, and rewrite (\ref{2.5}) as a linear state space system as
\begin{equation}
    \dot Y=AY+Bu
    \label{2.6}
\end{equation}
where $A=\begin{bmatrix}\mathbf{0}&I\\-M^{-1}K&-M^{-1}D\end{bmatrix},B=\begin{bmatrix}\mathbf{0}&\mathbf{0}\\M^{-1}&M^{-1}\end{bmatrix}$, in which $I\in \mathcal{R}^{6\times 6}$ denoting an identity matrix, $u=\left[f_h, f_r\right]^T\in\mathbb{R}^{12}$.  %
Next, we consider human and robot as two players and define their cost functions \cite{franceschiModelingAnalysisPHRI2023} as
\begin{equation}
    \begin{aligned}
        J_{h}=&\int_0^\infty\left[(Y-Y_{ref,h})^TQ_{h,h}(Y-Y_{ref,h})\right.\\
        &\left.+(Y-Y_{ref,r})^TQ_{h,r}(Y-Y_{ref,r})+f_hR_hf_h\right]dt\\
        J_{r}=&\int_0^\infty\left[(Y-Y_{ref,r})^TQ_{r,r}(Y-Y_{ref,r})\right.\\
        &\left.+(Y-Y_{ref,h})^TQ_{r,h}(Y-Y_{ref,h})+f_rR_rf_r\right]dt
    \end{aligned}
    \label{2.7}
\end{equation}
where $Q_{h,h}, Q_{h,r},Q_{r,h}, Q_{r,r}$ are the tracking error weights, and $R_h, R_r$ are the control input weights. Based on DCGT \cite{franceschiAdaptiveImpedanceController2022}, we define the shared cost function in cooperation as
\begin{equation}
    \begin{aligned}
    J_{c}&=\kappa J_{h}+(1-\kappa) J_{r}\\
    &=\int_{0}^{\infty}\left((Y-Y_{ref})^{T} Q_{\kappa } (Y-Y_{ref})+u^{T} R_{\kappa } u\right) dt
    \end{aligned}
    \label{2.8}
\end{equation}
where $Q_{\kappa }, R_{\kappa }$ are defined as $Q_{\kappa}=\kappa\left(Q_{h,h}+Q_{h,r}\right)+\left(1-\kappa\right)\left(Q_{r,h}+Q_{r,r}\right)$, $R_{\kappa}=\mathrm{diag}\{\kappa R_h,(1-\kappa)R_r\}$, $\kappa\in[0,1]$ is the role allocation coefficient that controls each cost's weight, and $Y_{ref}$ is a weighted composition of human and robot reference trajectories, $Y_{ref}=\hat{Y}$ when the predictions are accurate, can be obtained by combining (\ref{2.7}) and (\ref{2.8}) as
\begin{equation}
    Y_{ref}=Q_{\kappa }^{-1}(Y_{ref,h} Q_{h} + Y_{ref,r} Q_{r})
    \label{2.9}
\end{equation}
where $Q_{h}=\alpha Q_{h,h}+(1-\alpha)Q_{h,r}$, $Q_{r}=\alpha Q_{r,h}+(1-\alpha)Q_{r,r}$.

The solution to minimize (\ref{2.8}) is a classical LQR problem. According to (\ref{2.6}), the optimal control input is defined as
\begin{equation}
    u=-K^*(Y-Y_{\mathrm{ref}})
    \label{2.10}
\end{equation}
where $K^*=-R_{\kappa}^{-1}B^TP^*$, and we can obtain $P^*$ by solving the algebraic Riccati equation (ARE)
\begin{equation}
    PA+A^TP+Q_{\kappa}-PBR_{\kappa}^{-1}B^TP=0
    \label{2.11}
\end{equation}

It is evident that the robot assistance level can be controlled by adjusting $\kappa$. In this paper, we adjust $\kappa$ according to environmental factors and task requirements to achieve optimal robot behavior. Specifically, when the robot operates in a hazardous environment, we set $\kappa$ to a larger value, allowing the robot to prioritize human intentions more to avoid potential dangers. Conversely, in a safe environment, we set $\kappa$ smaller, enabling the robot to focus more on task requirements, thereby reducing disturbances that may arise from human involvement and environment. $\kappa$ is based on the human-applied force $f_h$, defined as
\begin{equation}
    \kappa = \frac{1}{1+e^{-\alpha \|f_h\|}}
    \label{2.12}
\end{equation}


By combining (\ref{2.9}) and (\ref{2.12}), we can obtain an adaptively adjusted reference trajectory and transform it into control inputs (\ref{2.10}), enabling role allocation in pHRC. Compared to existing methods\cite{franceschiLearningHumanMotion2023,maHumanRobotCollaboration2024,cremerModelFreeOnlineNeuroadaptive2020}, this method preserves the robot's autonomy in obstacle-free case through $\kappa$, while adaptively reducing human-robot discrepancies during obstacle avoidance, thereby enhancing the robot assistance level and ensuring efficient task completion.

\section{Experiments}


This section presents three sets of experiments to comprehensively evaluate the effectiveness of DTRT. The robotic platform is a UR5 robot, equipped with a Robotiq FT300S Force/Torque Sensor and a 2F-85 gripper. The data collection process, simulations, and comparison experiments are detailed in the accompanying multimedia submission.

\subsection{Estimation Performance in Simulation}




In this part, we compare estimation performance of DTRT with SOTA methods\cite{songRobotTrajectronTrajectory2024,franceschiLearningHumanMotion2023,9412190} on Traj100k\cite{songRobotTrajectronTrajectory2024} with a sampling frequency of 20 Hz. Traj100k contains 100,000 trajectories, of which 90,000 are used for training and 10,000 for testing. We employ observation length $T_{\mathrm{obs}}$ of 8 time steps and future horizon $T_{\mathrm{fut}}$ of 12 time steps for prediction. Human-guided module is not enabled in the simulation.

We utilize Average Displacement Error (ADE) and Final Displacement Error (FDE) as evaluation metrics. ADE measures the average L2 distance between all predictions and Ground Truth (GT) values over $(T_{\mathrm{now}},T_{\mathrm{now}}+T_{\mathrm{fut}}]$, while FDE is the L2 distance between the predicted final position and the ground truth value at $T_{\mathrm{now}}+T_{\mathrm{fut}}$. To calculate these metrics, we sampled the best-of-20 and most likely \cite{ivanovicTrajectronProbabilisticMultiAgent2019} trajectories from the predictions by each model.

\begin{table}[!t]
    \centering
    \caption{Comparison with SOTA methods on Traj100k}
    \begin{tabular}{cccccc}
    \hline
    \multirow{2}{*}{Method} & \multicolumn{2}{c}{Best-of-20(mm)} &  & \multicolumn{2}{c}{Most likely(mm)} \\ \cline{2-3} \cline{5-6} 
                                                      & ADE              & FDE       &        & ADE              & FDE              \\ \hline
    LSTM+FC\cite{franceschiLearningHumanMotion2023}   & -                & -         &        & 60.73            & 138.41            \\
    Transformer\cite{9412190}                         & -                & -         &        & 47.71            & 88.05            \\
    RT\cite{songRobotTrajectronTrajectory2024}        & 17.82            & 26.97     &        & 30.58            & 49.94            \\
    DTRT                                              & 14.51            & 21.56     &        & 24.21            & 38.26            \\
    \hline
    \end{tabular}
    \label{tab:4.1}
\end{table}

\begin{figure}[!t] 
    \centering
    \includegraphics[width=0.475\textwidth]{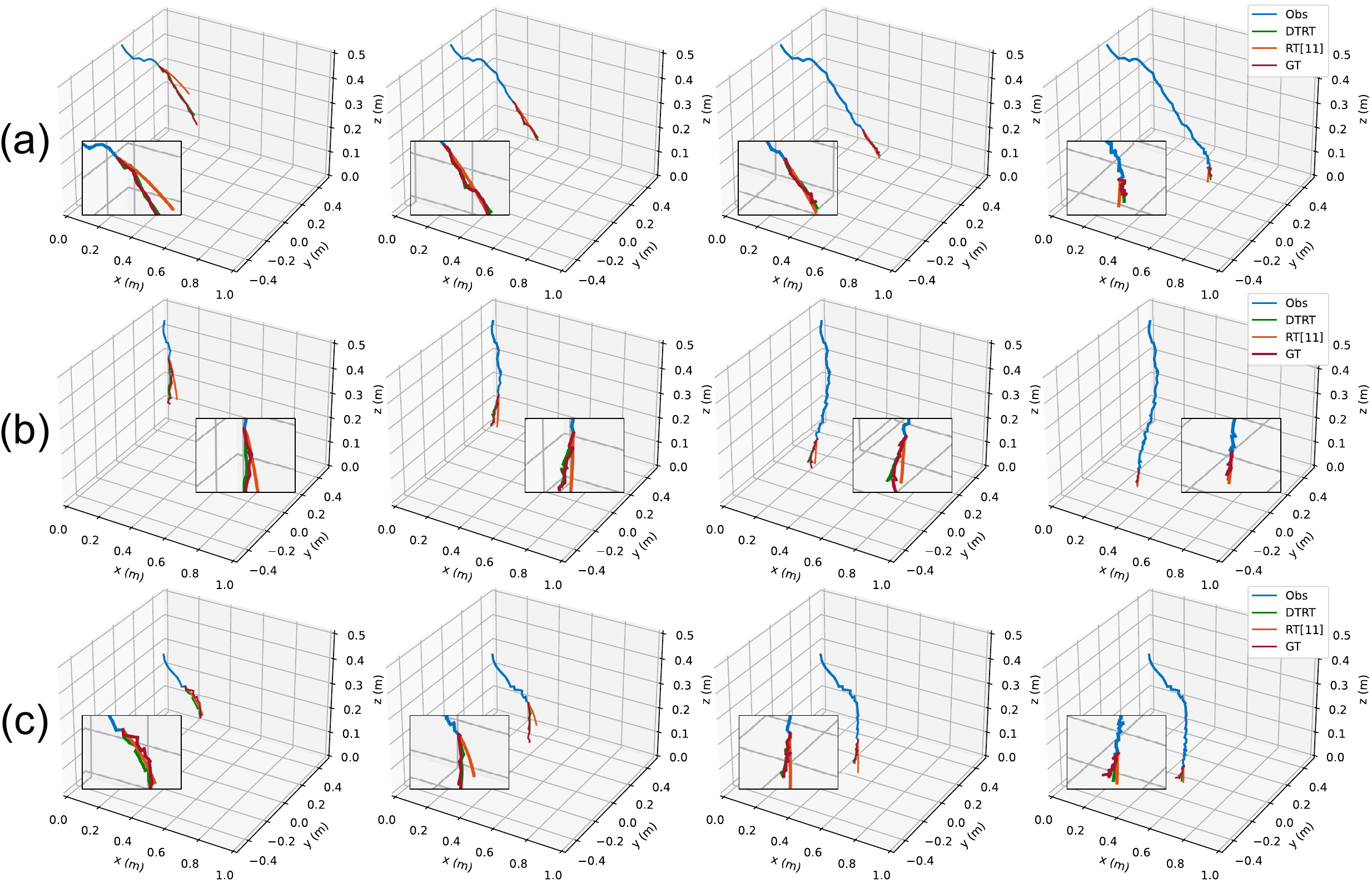}
    \caption{Visualization of predictions. The blue lines represent past trajectory, the red lines represent ground truth, the orange lines represent predictions of RT\cite{songRobotTrajectronTrajectory2024}, and the green lines represent prediction of DTRT. }
    \label{fig:2.3}
\end{figure}
\begin{figure}[!t] 
    \centering
    \includegraphics[width=0.465\textwidth]{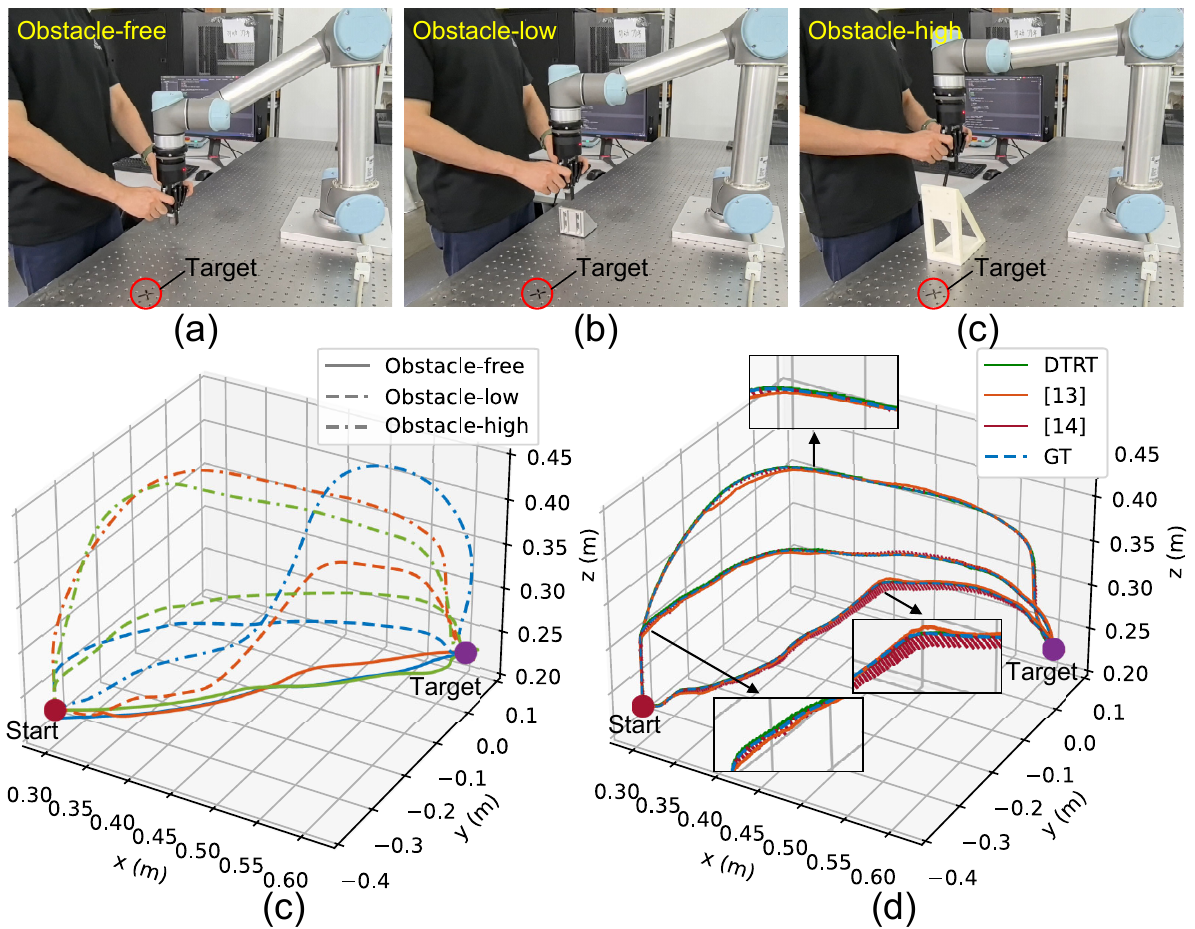}
    \caption{Data collection and predictions. (a)-(c) Data collection processes. (d) Examples of collected trajectories, where solid lines are obstacle-free case, dashed lines are low obstacle case, and dotted lines are high obstacle case. Each case contains 3 trajectories. (e) Visualization of predictions.}
    \label{fig:2.4}
\end{figure}

The results are shown in Table \ref{tab:4.1}. Since \cite{franceschiLearningHumanMotion2023, 9412190} are deterministic methods, we only report their most likely results. It is clear that the incorporation of Transformer enhances the prediction accuracy. Compared to RT, DTRT reduces ADE by 18.57\% and FDE by 20.06\% on best-of-20. On most likely, DTRT reduces ADE by 20.83\% and FDE by 23.39\%. However, due to the low sampling frequency of Traj100k, the prediction errors are relatively large across all models. Fig. \ref{fig:2.3} presents some examples of predictions generated by RT and DTRT. It can be seen that DTRT outperforms RT in majority of predictions, particularly when the trajectory deviates due to sudden changes in human intentions.

\subsection{DTRT Performance in PHRC with Hazardous}




In this part, we validate the effectiveness of DTRT in real-world pHRC and compare it with existing methods\cite{franceschiLearningHumanMotion2023, cremerModelFreeOnlineNeuroadaptive2020, maHumanRobotCollaboration2024}. In the experiment, robot and human move along a shared trajectory with an obstacle that the robot cannot actively avoid. Upon noticing the obstacle, human guides the robot away from danger, simulating intention changes. 

In data collection, we used UR5 in teach mode to gather robot trajectories in obstacle-free (Case 1) and trajectories/forces during human-guided obstacle avoidance (Case 2), as shown in Fig. \ref{fig:2.4}. The end-effector was moved from start pose to target. We used two obstacles of different heights, with their positions randomly set. We collected a total of 100 sets of data at 100 Hz. Among them, there are 40 sets for obstacle-free case and 60 sets for obstacle avoidance case. 

Based on experience, the impedance parameters (\ref{2.5}) are set as $M = \mathrm{diag}([10, 10, 10])$, $D = \mathrm{diag}([100, 100, 100])$, $K = \mathrm{diag}([200, 200, 200])$. The control parameters (\ref{2.7}) are set as $Q_{h,h}=Q_{r,r} = \mathrm{diag}([1, 1, 1, 0.0001, 0.0001, 0.0001])$, $Q_{h,r}=Q_{r,h} = 0^{6\times 6}$, $R_h = \mathrm{diag}([0.0005, 0.0005, 0.0005])$, $R_r = \mathrm{diag}([0.0001, 0.0001, 0.0001])$. The hyperparameter $\alpha $ (\ref{2.12}) is set as 0.3. Human intent estimator DTRT runs at 50 Hz, and the control frequency is set as 100 Hz. 


\begin{table}[!t]
    \centering
    \caption{Comparison with existing methods in pHRC}
    \begin{tabular}{cccccc}
    \hline
    Method & $e_{\mathrm{avg}}\ (\mathrm{mm})$ & $e_{\mathrm{max}}\ (\mathrm{mm})$ & $t_{\mathrm{avg}}\ (\mathrm{ms})$ \\\hline
    \cite{cremerModelFreeOnlineNeuroadaptive2020}     & 2.56            & 11.03        & 0.34   \\
    \cite{maHumanRobotCollaboration2024}              & 1.94            & 10.45        & 0.50   \\
    \cite{franceschiLearningHumanMotion2023}          & 3.79             & 17.63         & 1.48  \\
    DTRT                                              & 0.26             & 1.39         & 3.47   \\
    \hline
    \end{tabular}
    \label{tab:4.2}
\end{table}

\begin{figure}[!t] 
    \centering
    \includegraphics[width=0.475\textwidth]{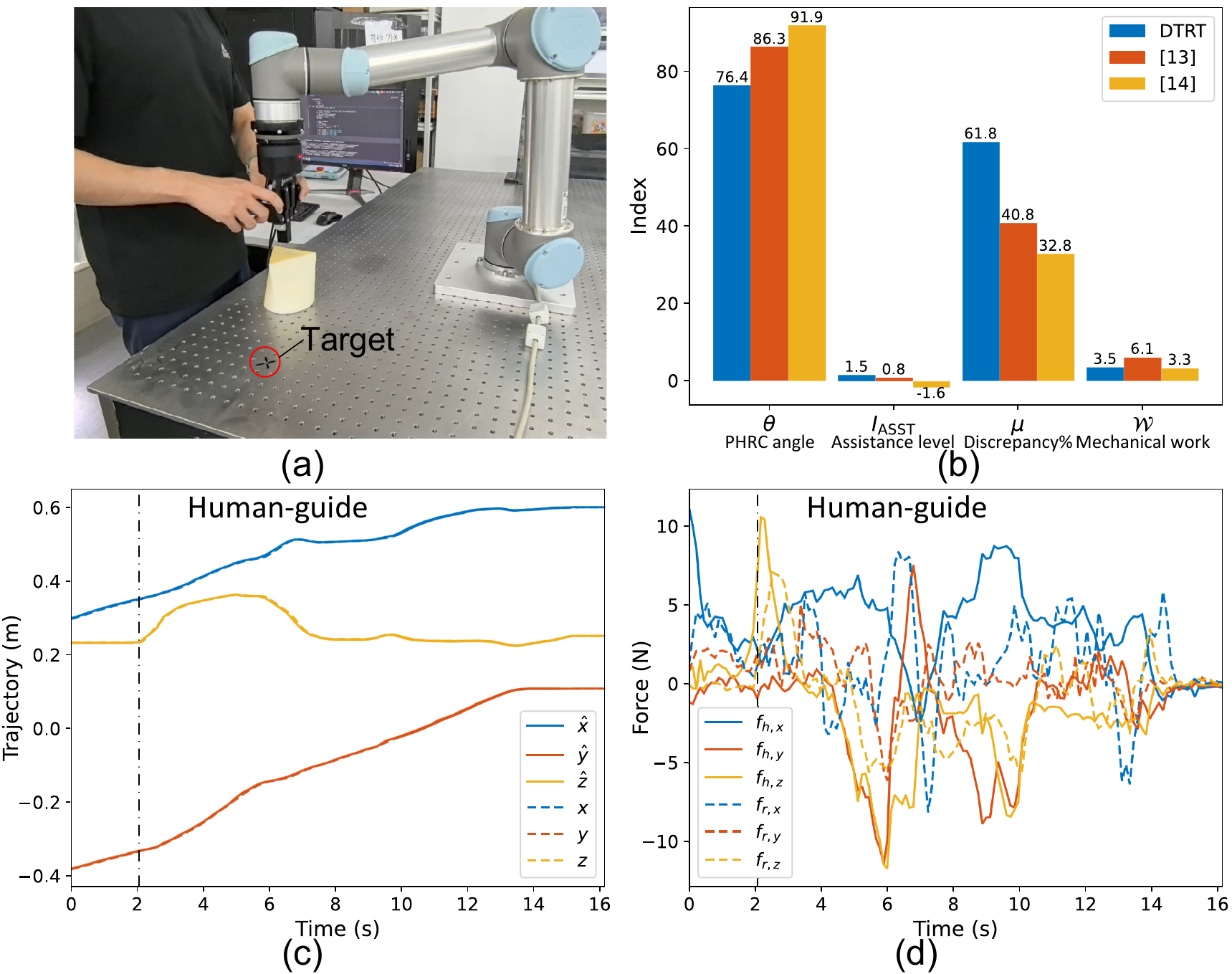}
    \caption{Process and results of pHRC. (a) PHRC process. (b) Comparison of DTRT with existing methods in terms of metrics. (c) DTRT predictions and actual robot trajectory. (d) Forces applied by human and robot.}
    \label{fig:2.5}
\end{figure}

{\noindent \bf Human Intent Estimation Performance.} First, we compare the prediction accuracy of DTRT with \cite{franceschiLearningHumanMotion2023, cremerModelFreeOnlineNeuroadaptive2020, maHumanRobotCollaboration2024} in simulation, where $\kappa=1$. Since the methods in \cite{cremerModelFreeOnlineNeuroadaptive2020,maHumanRobotCollaboration2024} only support short-term prediction, we use the average prediction error $e_{\mathrm{avg}}$ and max prediction error $e_{\mathrm{max}}$ as evaluation metrics. Meanwhile, to validate the real-time performance of these methods, we evaluate the average prediction time $t_{\mathrm{avg}}$. The results are shown in Table \ref{tab:4.2}. With the increased sampling frequency, the prediction accuracies of all models have improved. Furthermore, Fig. \ref{fig:2.4} (d) illustrates the prediction results of the above methods on three example trajectories. Clearly, DTRT also excels in trajectory prediction within the pHRC scenario, yielding results that closely match the ground truth, with $e_{\mathrm{avg}}=0.26\ \mathrm{mm}$ and $e_{\mathrm{max}}=1.39\ \mathrm{mm}$, significantly outperforming the existing methods. This substantial improvement enables the robot to accurately identify and estimate human intentions, thus reducing human-robot discrepancies, enhancing the robot assistance level, and providing support for collaborative operations. Regarding prediction efficiency, the average prediction time $t_{\mathrm{avg}}$ of DTRT is $3.47\ \mathrm{ms}$, which meets the requirements of pHRC.

{\noindent \bf Real-world PHRC Performance.} Next, we evaluate the pHRC performance of DTRT and existing methods \cite{franceschiLearningHumanMotion2023, cremerModelFreeOnlineNeuroadaptive2020, maHumanRobotCollaboration2024} in hazardous conditions, where $\kappa$ is adjusted according to (\ref{2.12}). Specifically, the evaluation used the following metrics: pHRC angle $\theta=\arccos\frac{f_r^Tf_h}{||f_r|||||f_h||}$, representing the average angle between $f_h$ and $f_r$; robot assistance level index $\mathcal{I} _{\mathrm{ASST}}=\frac{f_{r}^{T}f_{h}}{||f_{h}||}$, denoting the projection of $f_r$ onto $f_h$; human-robot assistance percentage $\mu=\frac{T_{\theta<90^{\circ}}}{T}$, indicating the proportion of time when $\theta < 90^{\circ}$ during motion; and human mechanical work $\mathcal{W}=\int_{0}^{T}f \mathrm{d}x$, a smaller value means the less human energy consumption in pHRC. 


\begin{figure}[!t] 
    \centering
    \includegraphics[width=0.45\textwidth]{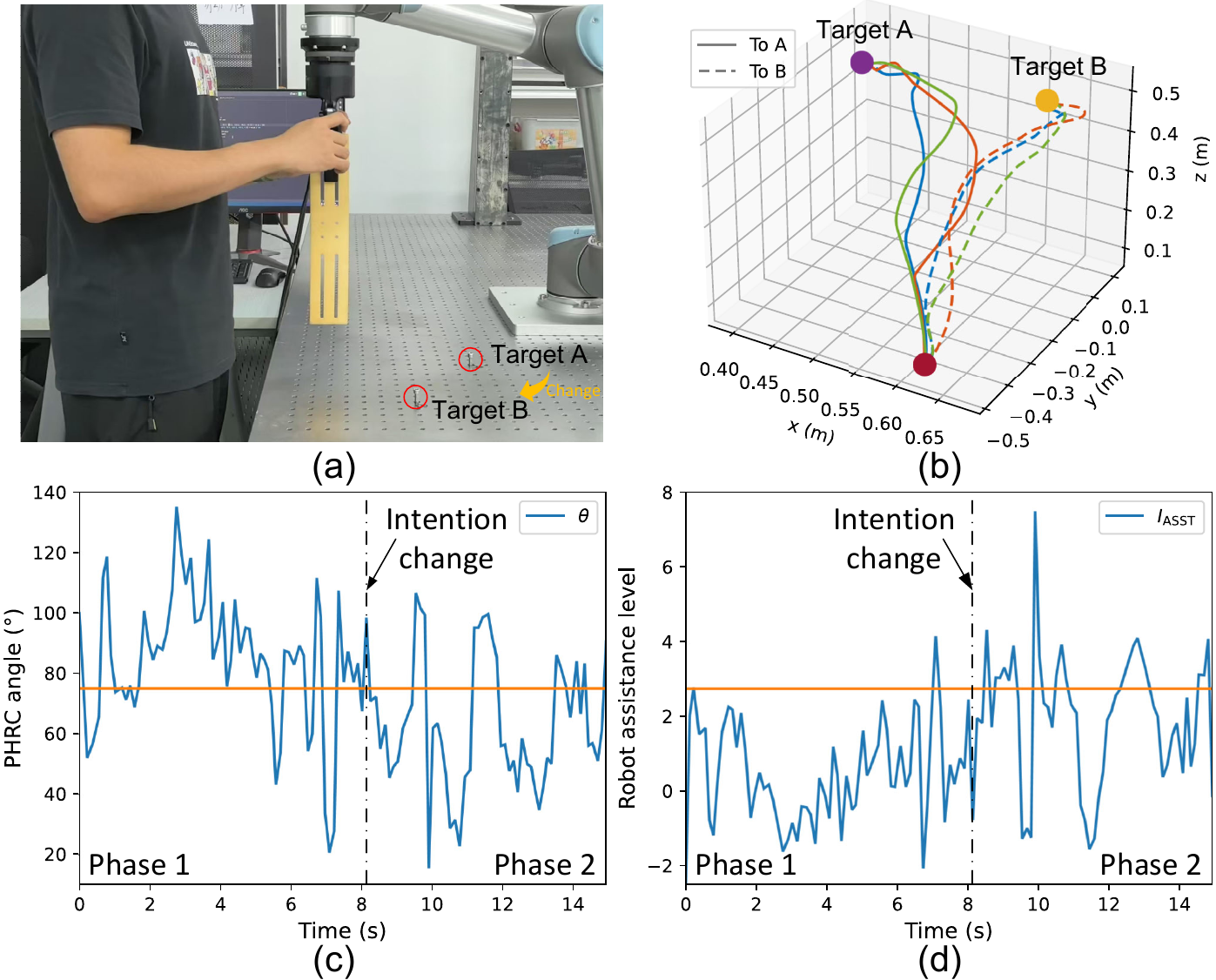}
    \caption{Results of co-transportation. (a) Experimental setup. (b) Examples of collected trajectories, where the targets of the solid lines are A, and the targets of the dotted lines are B. (c) PHRC angle $\theta$. (d) Robot assistance level index $\mathcal{I} _{\mathrm{ASST}}$.}
    \label{fig:2.6}
\end{figure}

In the experiment, we replace and randomly position the obstacle. Unlike \cite{maHumanRobotCollaboration2024} applying full-time human guidance, \cite{franceschiLearningHumanMotion2023} and DTRT switch human-robot roles when needed. Therefore, except for human mechanical work $\mathcal{W}$, the other metrics of \cite{franceschiLearningHumanMotion2023} and DTRT are only calculated during the human-guided phase. Fig. \ref{fig:2.5} shows the experimental results and compares DTRT with \cite{franceschiLearningHumanMotion2023, maHumanRobotCollaboration2024} based on the metrics. The results indicate that DTRT outperforms existing methods in pHRC scenarios. The average pHRC angle of \cite{franceschiLearningHumanMotion2023} is $91.9^\circ $, the average robot assistance level index is -1.6, and the human–robot system is in a state of mutual assistance for only 32.8\% of the time, implying that robot hinders human movement. And the average pHRC angle of \cite{maHumanRobotCollaboration2024} is $86.3^\circ $, the average robot assistance level index is 0.8, and the human–robot system is in a state of mutual assistance for 40.8\% of the time. While \cite{maHumanRobotCollaboration2024} demonstrates collaboration potential through its pHRC angle and assistance level, its fully human-guided method overlooks the robot's autonomy, resulting in greater human mechanical work. In comparison, DTRT achieves an average pHRC angle of $76.4^\circ $, an average robot assistance level index of 1.5, and the human–robot system is in a state of mutual assistance for 61.8\% of the time. Furthermore, the role allocation between human and robot effectively balances the robot's autonomy with human intention, resulting in only $3.5\ \mathrm{J}$ human mechanical work.

\subsection{Application Implementation and Validation}




In this part, we verify the real-world practicality of DTRT through co-transportation and drawing tasks.

First, in human-robot co-transportation, we simulate sudden change in human intention. Robot is tasked with transporting a plastic sheet to target A. At one point, human changes target to B and guides the robot. The experimental setup is shown in Fig. \ref{fig:2.6} (a). We collected 15 trajectories each of targets A and B, as shown in Fig. \ref{fig:2.6} (b). In Phase 1, robot takes the lead and operates autonomously. In Phase 2, human assumes control, with robot effectively supporting the task, reducing human effort. PHRC angle and robot assistance level in co-transportation are shown in Fig. \ref{fig:2.6} (c)-(d). It is evident that in Phase 1, where robot leads, average pHRC angle is above $90^\circ $, and assistance level remains around 0. After human intent changes (Phase 2), robot becomes follower and human leads, leading to the average pHRC angle below $90^\circ $ and an increase in assistance level.

\begin{figure}[!t] 
    \centering
    \includegraphics[width=0.475\textwidth]{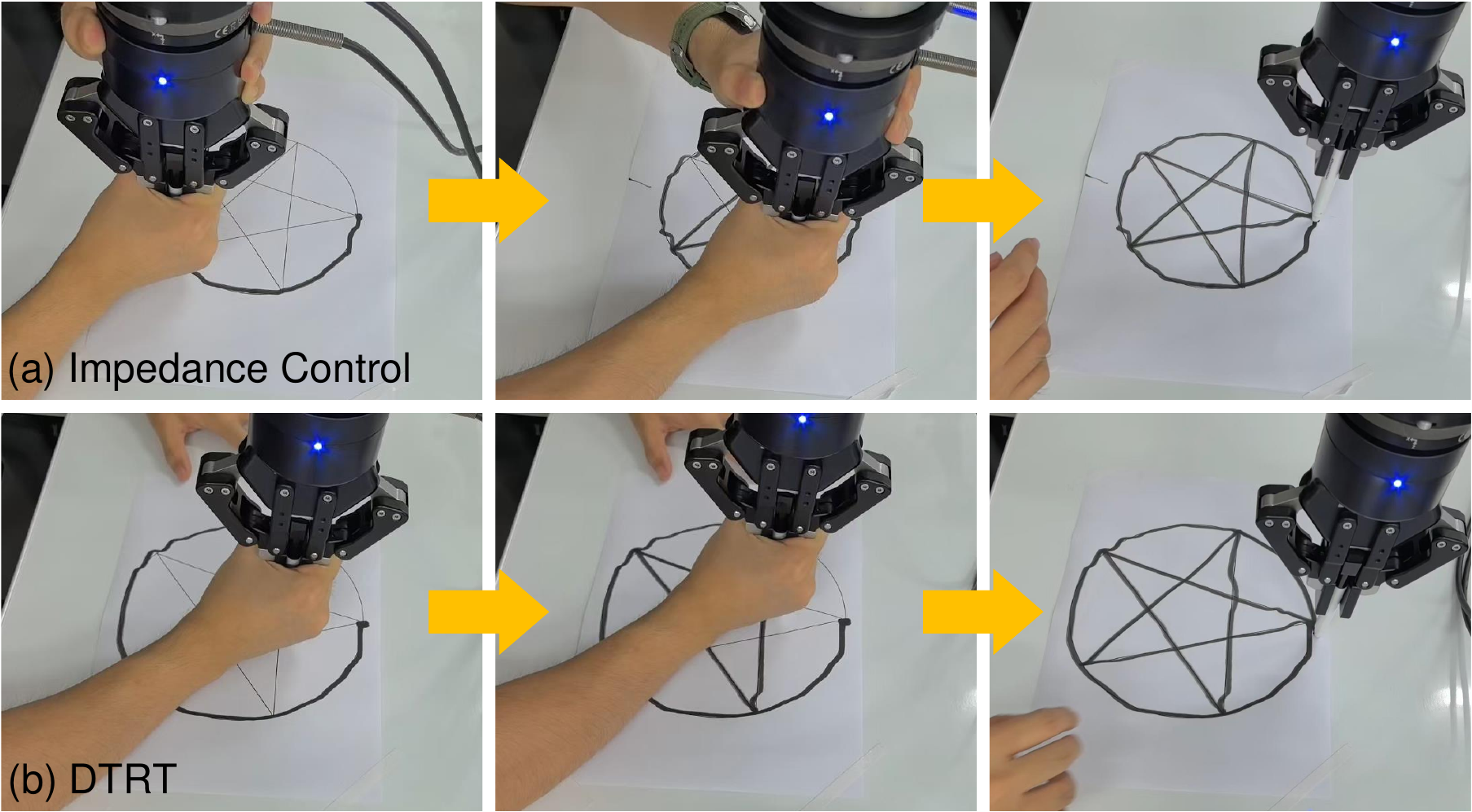}
    \caption{Processes of drawing. (a) Traditional impedance control. (b) DTRT.}
    \label{fig:2.7}
\end{figure}

Next, in drawing, human and robot collaboratively plot a circle. At one point, human decides to switch the intention to a pentagram, requiring robot to quickly adapt its behavior to assist in completing the drawing task. Fig. \ref{fig:2.7} shows the results of DTRT compared to traditional impedance control. Obviously, robot using DTRT demonstrates good autonomy when drawing circle and quickly transfer control to human after intention change. However, shortcomings in demonstrations cause an imperfect result. In traditional impedance control, two-handed operation is required to follow the trajectory, making it difficult to collaborate; whereas with DTRT, better drawing results can be achieved with only one hand.

Overall, DTRT accurately estimates changes in human intentions and rationally allocate roles of leader and follower, thereby reducing human-robot discrepancies and enhancing robot assistance level, making pHRC safe and efficient.

\section{Conclusion}


This work proposes DTRT, designed to accurately estimate human intentions and appropriately allocate roles in pHRC. DTRT employs a hierarchical architecture that effectively utilizes motion and force data to accurately predict future trajectories and rapidly adapt to changes in human intentions. Experiments demonstrate that DTRT offers significant advantages in prediction accuracy, effectively enhancing robot assistance level for pHRC, showcasing its great potential for applications in complex and hazard-prone environments. Future work will focus on integrating strengths of LSTM and Transformer, while considering robot dynamics uncertainties.

\bibliographystyle{IEEEtranBST/IEEEtran}
\bibliography{IEEEtranBST/IEEEabrv,reference}

\begin{thebibliography}{10}
\providecommand{\url}[1]{#1}
\csname url@rmstyle\endcsname
\providecommand{\newblock}{\relax}
\providecommand{\bibinfo}[2]{#2}
\providecommand\BIBentrySTDinterwordspacing{\spaceskip=0pt\relax}
\providecommand\BIBentryALTinterwordstretchfactor{4}
\providecommand\BIBentryALTinterwordspacing{\spaceskip=\fontdimen2\font plus
\BIBentryALTinterwordstretchfactor\fontdimen3\font minus
  \fontdimen4\font\relax}
\providecommand\BIBforeignlanguage[2]{{%
\expandafter\ifx\csname l@#1\endcsname\relax
\typeout{** WARNING: IEEEtran.bst: No hyphenation pattern has been}%
\typeout{** loaded for the language `#1'. Using the pattern for}%
\typeout{** the default language instead.}%
\else
\language=\csname l@#1\endcsname
\fi
#2}}

\bibitem{xingImpedanceLearningHumanGuided2023}
X.~Xing, E.~Burdet, W.~Si, C.~Yang, and Y.~Li, ``Impedance learning for
  human-guided robots in contact with unknown environments,'' \emph{IEEE
  Transactions on Robotics}, vol.~39, no.~5, pp. 3705--3721, 2023.

\bibitem{10309357}
C.~Shi, J.~Madera, H.~Boyea, and A.~M. Fey, ``Haptic guidance using a
  transformer-based surgeon-side trajectory prediction algorithm for
  robot-assisted surgical training,'' in \emph{2023 32nd IEEE International
  Conference on Robot and Human Interactive Communication (RO-MAN)}, 2023, pp.
  1942--1949.

\bibitem{yuAdaptiveConstrainedImpedanceControl2022}
X.~Yu, B.~Li, W.~He, Y.~Feng, L.~Cheng, and C.~Silvestre,
  ``Adaptive-constrained impedance control for human-robot co-transportation,''
  \emph{IEEE Transactions on Cybernetics}, vol.~52, no.~12, pp.
  13\,237--13\,249, 2022.

\bibitem{raviProactiveSafeHumanRobot2024}
P.~Ravi, W.~Zhuoyuan, N.~Yorie, and L.~Changliu, ``Towards proactive safe
  human-robot collaborations via data-efficient conditional behavior
  prediction,'' in \emph{2024 IEEE International Conference on Robotics and
  Automation ({{ICRA}})}.\hskip 1em plus 0.5em minus 0.4em\relax Yokohama,
  Japan: IEEE, 2024, pp. 12\,956--12\,963.

\bibitem{mengHierarchicalHumanMotion2024}
L.~Meng, L.~Yang, and E.~Zheng, ``Hierarchical human motion intention
  prediction for increasing efficacy of human-robot collaboration,'' \emph{IEEE
  Robotics and Automation Letters}, vol.~9, no.~9, pp. 7637--7644, 2024.

\bibitem{liuFollowForceHaptic2023}
Y.~Liu, R.~Leib, and D.~W. Franklin, ``Follow the force: Haptic communication
  enhances coordination in physical human-robot interaction when humans are
  followers,'' \emph{IEEE Robotics and Automation Letters}, vol.~8, no.~10, pp.
  6459--6466, 2023.

\bibitem{zhangOnlineHumanDynamic2023}
T.~Zhang, H.~Chu, and Y.~Zou, ``An online human dynamic arm strength perception
  method based on surface electromyography signals for human-robot
  collaboration,'' \emph{IEEE Transactions on Instrumentation and Measurement},
  vol.~72, pp. 1--14, 2023.

\bibitem{8616062}
L.~Roveda, N.~Castaman, S.~Ghidoni, P.~Franceschi, N.~Boscolo, E.~Pagello, and
  N.~Pedrocchi, ``Human-robot cooperative interaction control for the
  installation of heavy and bulky components,'' in \emph{2018 IEEE
  International Conference on Systems, Man, and Cybernetics (SMC)}, 2018, pp.
  339--344.

\bibitem{luoPhysicalHumanRobot2024}
J.~Luo, C.~Zhang, W.~Si, Y.~Jiang, C.~Yang, and C.~Zeng, ``A physical
  human-robot interaction framework for trajectory adaptation based on human
  motion prediction and adaptive impedance control,'' \emph{IEEE Transactions
  on Automation Science and Engineering}, pp. 1--12, 2024.

\bibitem{franceschiAdaptiveImpedanceController2022}
P.~Franceschi, N.~Pedrocchi, and M.~Beschi, ``Adaptive impedance controller for
  human-robot arbitration based on cooperative differential game theory,'' in
  \emph{2022 International Conference on Robotics and Automation (ICRA)}.\hskip
  1em plus 0.5em minus 0.4em\relax Philadelphia, PA, USA: IEEE, 2022, pp.
  7881--7887.

\bibitem{songRobotTrajectronTrajectory2024}
P.~Song, P.~Li, E.~Aertbelien, and R.~Detry, ``Robot trajectron: Trajectory
  prediction-based shared control for robot manipulation,'' in \emph{2024 IEEE
  International Conference on Robotics and Automation (ICRA)}.\hskip 1em plus
  0.5em minus 0.4em\relax Yokohama, Japan: IEEE, 2024.

\bibitem{cremerModelFreeOnlineNeuroadaptive2020}
S.~Cremer, S.~K. Das, I.~B. Wijayasinghe, D.~O. Popa, and F.~L. Lewis,
  ``Model-free online neuroadaptive controller with intent estimation for
  physical human-robot interaction,'' \emph{IEEE Transactions on Robotics},
  vol.~36, no.~1, pp. 240--253, 2020.

\bibitem{maHumanRobotCollaboration2024}
M.~Ma and L.~Cheng, ``A human-robot collaboration controller utilizing
  confidence for disagreement adjustment,'' \emph{IEEE Transactions on
  Robotics}, vol.~40, pp. 2081--2097, 2024.

\bibitem{franceschiLearningHumanMotion2023}
P.~Franceschi, F.~Bertini, F.~Braghin, L.~Roveda, N.~Pedrocchi, and M.~Beschi,
  ``Learning human motion intention for phri assistive control,'' in \emph{2023
  IEEE/RSJ International Conference on Intelligent Robots and Systems
  (IROS)}.\hskip 1em plus 0.5em minus 0.4em\relax Detroit, MI, USA: IEEE, 2023,
  pp. 7870--7877.

\bibitem{shao2024constraintaware}
Y.~Shao, T.~Li, S.~Keyvanian, P.~Chadhuari, V.~Kumar, and N.~Figueroa,
  ``Constraint-aware intent estimation for dynamic human-robot object
  co-manipulation,'' in \emph{Proceedings of Robotics: Science and Systems},
  Delft, Netherlands, July 2024.

\bibitem{sharifiImpedanceLearningBasedAdaptive2022}
M.~Sharifi, V.~Azimi, V.~K. Mushahwar, and M.~Tavakoli, ``Impedance
  learning-based adaptive control for human-robot interaction,'' \emph{IEEE
  Transactions on Control Systems Technology}, vol.~30, no.~4, pp. 1345--1358,
  2022.

\bibitem{pengNeuralControlHuman2024}
G.~Peng, C.~Yang, and C.~L.~P. Chen, ``Neural control for human-robot
  interaction with human motion intention estimation,'' \emph{IEEE Transactions
  on Industrial Electronics}, pp. 1--10, 2024.

\bibitem{wangRoleDynamicAssignment2024}
C.~Wang and J.~Zhao, ``Role dynamic assignment of human-robot collaboration
  based on target prediction and fuzzy inference,'' \emph{IEEE Transactions on
  Industrial Informatics}, vol.~20, no.~1, pp. 471--481, 2024.

\bibitem{xingDynamicMotionPrimitivesBased2024}
X.~Xing, K.~Maqsood, C.~Zeng, C.~Yang, S.~Yuan, and Y.~Li, ``Dynamic motion
  primitives-based trajectory learning for physical human-robot interaction
  force control,'' \emph{IEEE Transactions on Industrial Informatics}, vol.~20,
  no.~2, pp. 1675--1686, 2024.

\bibitem{Huang2005SharedNC}
C.~Huang, G.~S. Wasson, M.~Alwan, P.~Sheth, and A.~Ledoux, ``Shared
  navigational control and user intent detection in an intelligent walker,'' in
  \emph{AAAI Fall Symposium: Caring Machines}, 2005.

\bibitem{6054057}
K.~Wakita, J.~Huang, P.~Di, K.~Sekiyama, and T.~Fukuda,
  ``Human-walking-intention-based motion control of an omnidirectional-type
  cane robot,'' \emph{IEEE/ASME Transactions on Mechatronics}, vol.~18, no.~1,
  pp. 285--296, 2013.

\bibitem{6943678}
A.~Radmand, E.~Scheme, and K.~Englehart, ``A characterization of the effect of
  limb position on emg features to guide the development of effective
  prosthetic control schemes,'' in \emph{2014 36th Annual International
  Conference of the IEEE Engineering in Medicine and Biology Society}, 2014,
  pp. 662--667.

\bibitem{6683036}
D.~P. McMullen, G.~Hotson, K.~D. Katyal, B.~A. Wester, M.~S. Fifer, T.~G.
  McGee, A.~Harris, M.~S. Johannes, R.~J. Vogelstein, A.~D. Ravitz, W.~S.
  Anderson, N.~V. Thakor, and N.~E. Crone, ``Demonstration of a semi-autonomous
  hybrid brain–machine interface using human intracranial eeg, eye tracking,
  and computer vision to control a robotic upper limb prosthetic,'' \emph{IEEE
  Transactions on Neural Systems and Rehabilitation Engineering}, vol.~22,
  no.~4, pp. 784--796, 2014.

\bibitem{gaoHybridRecurrentNeural2023}
X.~Gao, L.~Yan, G.~Wang, and C.~Gerada, ``Hybrid recurrent neural network
  architecture-based intention recognition for human-robot collaboration,''
  \emph{IEEE Transactions on Cybernetics}, vol.~53, no.~3, pp. 1578--1586,
  2023.

\bibitem{dominguez-vidalExploringTransformersVisual2024}
J.~E. Dom{\'i}nguez-Vidal and A.~Sanfeliu, ``Exploring transformers and visual
  transformers for force prediction in human-robot collaborative transportation
  tasks,'' in \emph{2024 IEEE International Conference on Robotics and
  Automation ({{ICRA}})}.\hskip 1em plus 0.5em minus 0.4em\relax Yokohama,
  Japan: IEEE, 2024, pp. 3191--3197.

\bibitem{fuscoTransformerBasedPredictionHuman2024}
A.~Fusco, V.~Modugno, D.~Kanoulas, A.~Rizzo, and M.~Cognetti,
  ``Transformer-based prediction of human motions and contact forces for
  physical human-robot interaction,'' in \emph{2024 IEEE International
  Conference on Robotics and Automation (ICRA)}.\hskip 1em plus 0.5em minus
  0.4em\relax Yokohama, Japan: IEEE, 2024, pp. 3161--3167.

\bibitem{7097058}
Y.~Li, K.~P. Tee, W.~L. Chan, R.~Yan, Y.~Chua, and D.~K. Limbu, ``Continuous
  role adaptation for human–robot shared control,'' \emph{IEEE Transactions
  on Robotics}, vol.~31, no.~3, pp. 672--681, 2015.

\bibitem{xingFuzzyLogicBasedArbitration2024}
X.~Xing, W.~Li, S.~Yuan, and Y.~Li, ``Fuzzy logic-based arbitration for shared
  control in continuous human-robot collaboration,'' \emph{IEEE Transactions on
  Fuzzy Systems}, vol.~32, no.~7, pp. 3979--3991, 2024.

\bibitem{7139983}
Y.~Li, K.~P. Tee, W.~L. Chan, R.~Yan, Y.~Chua, and D.~K. Limbu, ``Role
  adaptation of human and robot in collaborative tasks,'' in \emph{2015 IEEE
  International Conference on Robotics and Automation (ICRA)}, 2015, pp.
  5602--5607.

\bibitem{7548305}
Y.~Li, K.~P. Tee, R.~Yan, W.~L. Chan, and Y.~Wu, ``A framework of human–robot
  coordination based on game theory and policy iteration,'' \emph{IEEE
  Transactions on Robotics}, vol.~32, no.~6, pp. 1408--1418, 2016.

\bibitem{7989336}
V.~Gabler, T.~Stahl, G.~Huber, O.~Oguz, and D.~Wollherr, ``A game-theoretic
  approach for adaptive action selection in close proximity
  human-robot-collaboration,'' in \emph{2017 IEEE International Conference on
  Robotics and Automation (ICRA)}, 2017, pp. 2897--2903.

\bibitem{liDifferentialGameTheory2019}
Y.~Li, G.~Carboni, F.~Gonzalez, D.~Campolo, and E.~Burdet, ``Differential game
  theory for versatile physical human-robot interaction,'' \emph{Nature Machine
  Intelligence}, vol.~1, no.~1, pp. 36--43, 2019.

\bibitem{franceschiHumanRobotRole2023}
P.~Franceschi, N.~Pedrocchi, and M.~Beschi, ``Human-robot role arbitration via
  differential game theory,'' \emph{IEEE Transactions on Automation Science and
  Engineering}, pp. 1--16, 2023.

\bibitem{9412190}
F.~Giuliari, I.~Hasan, M.~Cristani, and F.~Galasso, ``Transformer networks for
  trajectory forecasting,'' in \emph{2020 25th international conference on
  pattern recognition (ICPR)}, 2021, pp. 10\,335--10\,342.

\bibitem{franceschiModelingAnalysisPHRI2023}
P.~Franceschi, M.~Beschi, N.~Pedrocchi, and A.~Valente, ``Modeling and analysis
  of {{pHRI}} with differential game theory,'' in \emph{2023 21st International
  Conference on Advanced Robotics ({{ICAR}})}.\hskip 1em plus 0.5em minus
  0.4em\relax Abu Dhabi, United Arab Emirates: IEEE, 2023, pp. 277--284.

\bibitem{ivanovicTrajectronProbabilisticMultiAgent2019}
B.~Ivanovic and M.~Pavone, ``The trajectron: Probabilistic multi-agent
  trajectory modeling with dynamic spatiotemporal graphs,'' in \emph{2019
  IEEE/CVF International Conference on Computer Vision (ICCV)}.\hskip 1em plus
  0.5em minus 0.4em\relax Seoul, Korea (South): IEEE, 2019, pp. 2375--2384.

\end{thebibliography}


\end{document}